\documentclass{llncs}

\usepackage{times}
\usepackage{color}
\usepackage{array,booktabs,multirow}
\usepackage{amsmath,amssymb}
\usepackage{latexsym}
\usepackage{amsfonts}
\usepackage{xspace}
\usepackage{graphics}
\usepackage{multirow}
\usepackage{graphicx}
\usepackage{todonotes}
\usepackage[all]{xy}
\usepackage{tikz}
\usepackage{csquotes}
\usepackage{paralist}
\usetikzlibrary{shapes.geometric,shapes.misc, positioning}
\usetikzlibrary{decorations.pathmorphing}

\newcommand{\prob}{P}
\newcommand{\args}{\ensuremath{\mathcal{A}}}
\newcommand{\attacks}{\ensuremath{\mathcal{R}}}
\newcommand{\supports}{\ensuremath{\mathcal{S}}}
\newcommand{\probDists}{\ensuremath{\mathcal{P}_\args}}
\newcommand{\probLabs}{\ensuremath{\mathcal{L}_\args}}

\newcommand{\sat}{\ensuremath{\mathrm{Sat}}}
\newcommand{\satp}{\ensuremath{\sat_\Pi}}
\newcommand{\satl}{\ensuremath{\sat_\Lambda}}
\newcommand{\constraints}{\ensuremath{\mathcal{C}_\args}}

\newcommand{\lsDistance}{\ensuremath{d^2_{\mathrm{At}}}}
\newcommand{\lsFirstUpdate}{\ensuremath{u_{\mathrm{At}}}}
\newcommand{\lsUpdate}{\ensuremath{\updateOp^2_{\mathrm{At}}}}
\newcommand{\lsLabellingUpdate}{\ensuremath{LU^2_{\lambda}}}
\newcommand{\updateOp}{\ensuremath{\mathcal{U}}}

\definecolor{darkgreen}{rgb}{0.09, 0.45, 0.27}

\def\showProofs{1}

\title{Polynomial-time Updates of Epistemic States in a Fragment of Probabilistic Epistemic Argumentation
\ifnum\showProofs=1
(Technical Report)
\fi
}

\author{Nico Potyka\inst{1} \and Sylwia Polberg\inst{2} \and Anthony Hunter\inst{3} }
\institute{Institute of Cognitive Science, University of Osnabr\"{u}ck, Germany,\\
           \and School of Computer Science and Informatics, Cardiff University, UK\\
          \and Department of Computer Science, University College London, UK }

\begin{document}

\maketitle


\begin{abstract}
Probabilistic epistemic argumentation allows for reasoning about argumentation problems
in a way that is well founded by probability theory. Epistemic states are represented 
by probability functions over possible worlds and can be adjusted to new beliefs using 
update operators.  While the use of probability functions puts this approach on a 
solid foundational basis, it also causes computational challenges 
as the amount of data to process depends exponentially on the number of arguments. 
This leads to bottlenecks in applications such as modelling opponent's beliefs 
for persuasion dialogues.  
We show how update operators over probability functions 
can be related to update operators over much more compact representations
that allow polynomial-time updates.
We discuss the cognitive and probabilistic-logical plausibility of this approach
and demonstrate its applicability in computational persuasion.
\end{abstract}

\section{Introduction}

Probabilistic epistemic argumentation \cite{thimm2012probabilistic,hunter2013probabilistic,HunterT16,HunterPT2018Arxiv} is an extension of Dung's classical argumentation framework \cite{dung1995acceptability}. 
While the original framework allows only for talking about attacks and accepting
or rejecting arguments, probabilistic epistemic argumentation also allows more general relationships between
arguments like support \cite{cayrol2013bipolarity,polberg2014revisiting,CohenGGS14} and allows expressing
more fine-grained beliefs by means of probabilities. Recent experiments give empirical evidence that these extensions are, in particular, beneficial when it comes to modelling human decision making \cite{polberg2018empirical}. One large application area of probabilistic epistemic argumentation is computational persuasion
\cite{Hunter15ijcai,Hunter16comma}. Computational persuasion aims at convincing the user of a persuasion goal such as giving up
bad habits or living a healthier lifestyle. In order to derive persuasion strategies autonomously, we require 
a user model that represents the user's beliefs and simulates belief changes when new
arguments are presented to the user. The user's epistemic state can be represented by a probability function
and different update operators have been studied that can be used to adapt the current beliefs \cite{Hunter15ijcai,HunterPotyka17,HunterPolbergPotyka18}.

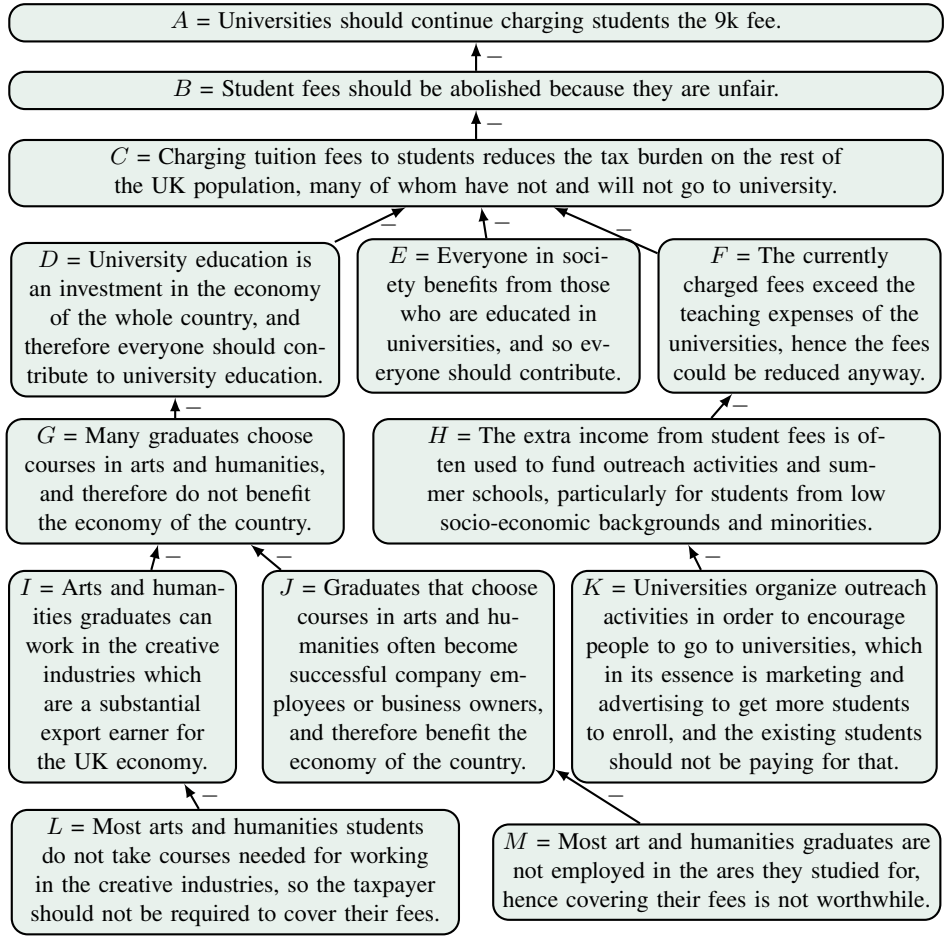
\begin{figure}[t!]
\centering
\begin{tikzpicture}[->,>=latex,thick, arg/.style={draw,text centered,
shape=rectangle, rounded corners=6pt,
fill=darkgreen!10,font=\small}]
\node[arg] (a1) [text width=\textwidth] at (0,2) {$A$ = Universities should continue charging students the 9k fee.};

\node[arg] (a2) [text width=\textwidth] at (0,1.1)  {$B$ = Student fees should be abolished because they are unfair.};

\node[arg] (a3) [text width=\textwidth] at (0,0)  {$C$ = Charging tuition fees to students reduces the tax burden on the rest of the UK population, many of whom have not and will not go to university.};

\node[arg] (a4a) [text width=0.34\textwidth] at (-4.0,-1.95)  {$D$ = University education is an investment in the economy of the whole country, and therefore everyone should contribute to university education.};

\node[arg] (a4b) [text width=0.29\textwidth] at (0.3,-1.9)  {$E$ = Everyone in society benefits from those who are educated in universities, and so everyone should contribute.};

\node[arg] (a4c) [text width=0.29\textwidth] at (4.3,-1.9)  {$F$ = The currently charged fees exceed the teaching expenses of the universities, hence the fees could be reduced anyway.};

\node[arg] (a5a) [text width=0.35	\textwidth] at (-4,-4.1)  {$G$ = Many graduates choose courses in arts and humanities, and therefore do not benefit the economy of the country.};
\node[arg] (a5b) [text width=0.6\textwidth] at (2.4,-4.1)  {$H$ = The extra income from student fees is often used to fund outreach activities and summer schools, particularly for students from low socio-economic backgrounds and minorities.};

\node[arg] (a6a) [text width=0.23\textwidth] at (-4.7,-6.7)  {$I$ = Arts and humanities graduates can work in the creative industries which are a substantial export earner for the UK economy.};

\node[arg] (a6b) [text width=0.3\textwidth] at (-0.9,-6.7)  {$J$ = Graduates that choose courses in arts and humanities often become successful company employees or business owners, and therefore benefit the economy of the country.};

\node[arg] (a6c) [text width=0.38\textwidth] at (3.7,-6.7)  {$K$ = Universities organize outreach activities in order to encourage people to go to universities, which in its essence is marketing and advertising to get more students to enroll, and the existing students should not be paying for that.};

\node[arg] (a7a) [text width=0.47\textwidth] at (-3.2,-9.3)  {$L$ = Most arts and humanities students do not take courses needed for working in the creative industries, so the taxpayer should not be required to cover their fees.};
\node[arg] (a7b) [text width=0.47\textwidth] at (3.2,-9.3)  {$M$ = Most art and humanities graduates are not employed in the ares they studied for, hence covering their fees is not worthwhile.};   

\path	(a2) edge node[right] {$-$}(a1);
\path	(a3) edge node[right] {$-$}(a2);
\path	(a4a) edge node[right] {$-$}(a3);
\path	(a4b) edge node[right] {$-$}(a3);
\path	(a4c) edge node[right] {$-$}(a3);
 
\path	(a5a) edge node[right] {$-$}(a4a); 
\path	(a5b) edge node[right] {$-$}(a4c);

\path	(a6a) edge node[right] {$-$}(a5a); 
\path	(a6b) edge node[right] {$-$}(a5a); 
\path	(a6c) edge node[right] {$-$}(a5b); 

\path	(a7a) edge node[right] {$-$}(a6a); 
\path	(a7b) edge node[right] {$-$}(a6b); 
\end{tikzpicture} 
\caption{\label{fig:dialogue2} Study fee dialogue. }
\end{figure}

Probability theory provides a strong foundational basis for probabilistic epistemic argumentation,
but also comes with computational limitations. Without further assumptions, probability
functions grow exponentially with the number of arguments. However, sometimes we are only 
interested in atomic beliefs in arguments, so that the full power of probability functions may
not be required. For instance, we can consider the graph depicted in Figure \ref{fig:dialogue2} 
induced by a dialogue between an automated dialogue system and a human participant that
occurred in the empirical study considered in \cite{studylink}. There are various constraints that could be attached to such a graph, as we will discuss further in Section \ref{sec:example}. 
For instance, we could use postulates from the classical epistemic approach \cite{thimm2012probabilistic,hunter2013probabilistic} 
such as \textit{coherence}, 
which bounds the belief in an argument based on the belief of its attacker. 
Formally, in our scenario, for every argument-attacker pair $X$ and $Y$ this would create a constraint
 of the form 
$\pi(X)+\pi(Y) \leq 1$, 
where $\pi(\alpha)$ should be read as the probability of $\alpha$. 

We observe that the aforementioned formulas operate on probabilities of single arguments rather than on complex logical expressions. 
Consequently, the detailed information contained in a full probability function can be seen as excessive.  
In such a situation, probability functions can sometimes be replaced by
probability labellings that assign probabilities to arguments directly without changing the
semantics \cite{potyka2019fragment}. 
In our case, this would decrease the number of probabilities that need to be processed from 8,192 (i.e. $2^{13}$) to 13, which has obvious computational benefits.

In this paper, we are interested in the relationship between 
epistemic states represented by probability functions and those represented by probability labellings.
Formally, probability labellings can be related
to equivalence classes of probability functions that assign the same atomic beliefs to arguments \cite{potyka2019fragment}. 
In order to  establish an interesting relationship, update operators must respect this equivalence 
relation. We define such an operator in Section 3 and show in Section 4 that it satisfies our desiderata.
In particular, updates  can be computed in polynomial time
In this approach, epistemic states correspond to sets of probability functions that satisfy the 
same atomic beliefs and updates are performed by satisfying the new beliefs while minimizing
the required changes. We will argue that this approach is not only computationally 
attractive, but can also result in cognitively more plausible updates. We illustrate our method 
with an application in computational persuasion in Section 5.
\ifnum\showProofs=0
All proofs for the results in this submission can be found in the technical report \todo{add reference}.
\fi
\section{Basics}
\label{sec_basics}

We consider \emph{bipolar argumentation frameworks (BAFs)} $(\args, \attacks, \supports)$
consisting of a set of arguments $\args$, an attack relation $\attacks \subseteq \args \times \args$ and a 
support relation $\supports \subseteq \args \times \args$.
$\Omega = \{w \mid w \subseteq \args\}$ denotes the set of \emph{possible worlds}.
 Intuitively, each
$w \in \Omega$ contains the arguments that are accepted in a particular state of the world. 
We represent beliefs by probability
functions $\prob: \Omega \rightarrow [0,1]$ such that $\sum_{w \in \Omega} \prob(w) = 1$.
$\probDists$ denotes the set of all probability functions over $\args$.
The probability of an argument $A \in \args$ under $\prob$
is defined by adding the probabilities of all worlds in which $A$ is accepted, that is, 
$\prob(A) = \sum_{w \in \Omega, A \in w} \prob(w)$. $\prob(A)$ can be understood as a degree of belief, 
where $\prob(A) = 1$ means complete acceptance and $\prob(A)=0$ means complete rejection\footnote{Note that $\prob(A)$ denotes the probability of argument $A$  (the sum of probabilities 
of all possible worlds that accept $A$), while $\prob(\{A\})$ denotes the probability of the possible world $\{A\}$.}.

The epistemic probabilistic argumentation approach developed in \cite{thimm2012probabilistic,hunter2013probabilistic,HunterT16,HunterPT2018Arxiv} defines semantics of attack and support relations
by means of constraints over probability functions. Some constraints can be automatically derived 
from the relations between arguments. For example, the \emph{coherence constraint} demands that if $A$ attacks $B$,
we must have $\prob(B) \leq 1 - \prob(A)$, that is, the belief in an attacked argument $B$ is bounded from
above by the belief in an attacker $A$. However, it is also possible to design individual constraints
manually. For example, if $B$ is attacked by three related arguments $A_1, A_2, A_3$, we may want to
bound the belief in $B$ by the average belief in these attackers via  $\prob(B) \leq 1 - \frac{1}{3}\sum_{i=1}^3 \prob(A_i)$.
To allow this flexibility, a general constraint language has been introduced in \cite{HunterPT2018Arxiv,HunterPolbergPotyka18}.
We will focus on the fragment of \emph{linear atomic constraints} here because it is sufficiently
expressive for most of the constraints considered in \cite{thimm2012probabilistic,hunter2013probabilistic,HunterT16} and sometimes allows polynomial-time computations \cite{potyka2019fragment}. 

Formally, 
a \emph{linear atomic constraint} over a set of arguments $\args$ is an expression of the form 
$\sum_{i=1}^n c_i \cdot \pi(A_i) \leq c_0$,
where $A_i \in \args$ and $c_i \in \mathbb{Q}$. $\pi$ is just a syntactic symbol that can
be read as \enquote{the probability of}.
We let $\constraints$ denote the set of all linear atomic constraints over $\args$.
A probability function $\prob$ \emph{satisfies} such a linear atomic constraint iff
$\sum_{i=1}^n c_i \cdot \prob(A_i) \leq c_0$.
$\prob$ satisfies a set of linear atomic constraints $C$, denoted as $\prob \models C$, iff it satisfies all $l \in C$.
In this case, we call $C$ \emph{satisfiable}.
We let $\satp(C) = \{P \in \probDists \mid P \models C\}$ denote the set of all probability functions that satisfy $C$.
We call sets of constraints $C_1, C_2$ equivalent and write 
$C_1 \equiv C_2$ iff they are satisfied by the same
probability functions, that is, $\satp(C_1) = \satp(C_2)$

Note that constraints with $\geq$ and $=$ can be expressed as well in our language. 
For $\geq$, just note that 
$\sum_{i=1}^n c_i \cdot \pi(A_i) \leq c_0$ is equivalent to
$\sum_{i=1}^n -c_i \cdot \pi(A_i) \geq -c_0$.
For $=$, note that 
$\sum_{i=1}^n c_i \cdot \pi(A_i) \leq c_0$ and
$\sum_{i=1}^n c_i \cdot \pi(A_i) \geq c_0$ together are equivalent to
$\sum_{i=1}^n c_i \cdot \pi(A_i) = c_0$.
In particular, we can express probability assignments of the form $\pi(A) = p$
or probability bounds of the form $l \leq \pi(A) \leq u$.

Now assume that we are given an epistemic state represented as a probability function
$\prob \in \probDists$. Given some new evidence represented as a set of linear atomic constraints (and possibly some existing 
constraints that we want to preserve),
we want to update $\prob$. 
To this end, different update operators have been studied in \cite{Hunter15ijcai,HunterPotyka17,HunterPolbergPotyka18}.
Here, we are interested in update operators of the following type.
\begin{definition}[Epistemic Update Operator]
\label{def_ep_update_function}
An \emph{epistemic update operator} is a function $\updateOp: \probDists \times \constraints  \rightarrow \probDists \cup \{\bot\}$ 
that satisfies the following properties:
\begin{itemize}
	\item \textbf{Success:} If $C \subseteq \constraints$ is satisfiable, then
	 $\updateOp(\prob, C) \in \satp(C)$.
	\item \textbf{Failure:} If  $C \subseteq \constraints$ is not satisfiable, then
	 $\updateOp(\prob, C) = \bot$.
	\item \textbf{Representation Invariance:} If $C_1 \equiv C_2$, then $\updateOp(\prob, C_1) = \updateOp(\prob,C_2)$.  
	\item \textbf{Idempotence:} If $C \subseteq \constraints$ is satisfiable, then $\updateOp(\updateOp(\prob, C), C) = \updateOp(\prob, C)$. 
\end{itemize} 
\end{definition}
Success and failure guarantee a well-defined update. That is, if the constraints are satisfiable,
the update operator will return a new epistemic state that satisfies the constraints.
If the constraints are not satisfiable, $\bot$ will be returned to indicate an inconsistency. 
Representation invariance guarantees that the result is independent of the syntactic representation 
of the evidence. Finally idempotence guarantees that applying the same update twice does not change
the outcome. 

\section{The Two-stage Least-squares Update Operator}

Several update operators in \cite{HunterPotyka17,HunterPolbergPotyka18} are based on the idea of
satisfying new evidence by changing the current epistemic state in a minimal way. 
The distance between two probability functions is
determined by looking at the probabilities that they assign to possible worlds. 
For example, one can use the least-squares distance $d_2(\prob,\prob') = \sum_{w \in \Omega} (\prob(w) - \prob'(w))^2$
or the KL-divergence $d_{KL}(\prob,\prob') = \sum_{w \in \Omega} \prob(w) \cdot \log \frac{\prob(w)}{\prob'(w)}$. 
While this makes perfect sense from a probability-theoretical point of view, 
the resulting belief changes may be intuitively implausible. 

\begin{table}[tb]
	\centering
		\begin{tabular}{lllllllllll}
		$w$ & $\prob_1 \qquad$ & $\prob_2 \qquad$ & $\prob_3 \qquad$ & $\prob_4 \qquad$ & $\prob_5 \qquad$  & $\prob_6 \qquad $ & 
		$\prob_7 \qquad $ & $\prob_{8} \qquad$ &  $\prob_{9} \qquad$  &  $\prob_{10} \quad$  \\
		\hline
		$\emptyset$ & $0.1$  & $0$   & $0$    & $0.4$  & $0.45$  
		            & $0.26$ & $0$   & $0.3$  & $0.15$ & $0.35$  \\
		$\{A\}$     & $0.2$  & $0.4$ & $0.33$ & $0.05$ & $0$  
		            & $0.19$ & $0.3$   & $0$    & $0.15$ & $0$  \\
		$\{B\}$     & $0.3$  & $0$   & $0$    & $0.15$ & $0.1$    
		            & $0.29$ & $0$   & $0.1$  & $0.35$ & $0.15$  \\
		$\{A, B\} \quad$ & $0.4$&$0.6$&$0.67$& $0.4$  & $0.45$ 
		            & $0.26$ & $0.7$   & $0.6$ & $0.35$ & $0.5$     
		\end{tabular}
	\caption{Some probability functions over possible worlds used in Example \ref{example_implausible_update}.}
	\label{tab:exampleAtomicUpdates}
\end{table}

\begin{example}
\label{example_implausible_update}
Consider a BAF $(\{A,B\}, \emptyset, \emptyset)$ with two unrelated arguments $A, B$.
Suppose our current epistemic state is $\prob_1$ as defined in Table \ref{tab:exampleAtomicUpdates}.
Then we have $\prob_1(A) = 0.6$ and $\prob_1(B) = 0.7$.
Now suppose that we 
want to update the belief in $A$ to $1$. 
A distance-minimizing update w.r.t. $d_2$ (i.e. update returning a probability distribution satisfying $\pi(A) = 1$ that is minimally different from $\prob_1$ w.r.t. $d_2$) yields the new epistemic state $\prob_2$ from Table \ref{tab:exampleAtomicUpdates}. 
Now we have $\prob_2(A) = 1$ as desired. 
However, we also have $\prob_2(B) = 0.6 < 0.7 = \prob_1(B)$.
Similarly, updating with respect to $d_{KL}$
yields $\prob_3$ from Table \ref{tab:exampleAtomicUpdates} with $\prob_3(B) = \frac{2}{3} < 0.7 = \prob_1(B)$.
This behaviour is rather counterintuitive in this context, since $A$ and $B$ are completely 
unrelated. Therefore, we should have $\prob_1(B) = \prob_2(B) = \prob_3(B)$.
\end{example}
In order to bring our model closer to humans' intuition, a two-stage minimization process has been
proposed in \cite{HunterPolbergPotyka18}. In stage 1, we identify all probability distributions 
that minimize an atomic distance measure. Instead of comparing probability functions elementwise on possible worlds, 
atomic distance measures compare probability functions only based on the probabilities
that they assign to arguments \cite{HunterPotyka17}.
We consider a quadratic variant here that will allow us to compute some updates in polynomial time.
\begin{definition}[Atomic Least-squares Distance (ALS)]
The {\bf  ALS distance measure} is defined as 
$\lsDistance(\prob,\prob') = \sum_{A \in \args} (\prob(A) - \prob'(A))^2$ for all $\prob, \prob' \in \probDists$.
\end{definition}
To begin with, we use the ALS distance to define a naive update operator which does not satisfy  
our desiderate from Definition \ref{def_ep_update_function} yet. 
\begin{definition}[Naive Least-squares Update Operator]
The \emph{naive LS update operator} $\lsFirstUpdate: \probDists \times \constraints  \rightarrow 2^{\probDists}$ is defined by
$
\lsFirstUpdate(\prob,C) = \arg \min_{\prob' \in \satp(C)} \lsDistance(\prob, \prob').
$
\end{definition}
$\lsFirstUpdate$ yields those probability functions that satisfy $C$ and minimize the ALS
distance to $\prob$. However, there is not necessarily a unique solution.
\begin{example}
\label{example_atomic_update_not_unique}
Consider $\prob_1$ from Table \ref{tab:exampleAtomicUpdates}.
Suppose we recognize a conflict between $A$ and $B$ and want to update with
the constraint $l_1: \pi(A) + \pi(B) \leq 1$.
We have $\prob_1(A) = 0.6$ and $\prob_1(B) = 0.7$. The cheapest way to satisfy
the constraint with respect to the ALS distance is to decrease both
probabilities by $0.15$. That is, a solution $\prob'$ must satisfy
$\prob'(A) = \prob'(\{A\}) + \prob'(\{A,B\}) = 0.45$ and
$\prob'(B) = \prob'(\{B\}) + \prob'(\{A,B\}) = 0.55$. 
$\prob_4$ and $\prob_5$  from Table \ref{tab:exampleAtomicUpdates} show two minimal
solutions from the set $\lsFirstUpdate(\prob_1, \{l_1\})$.
\end{example}
The second stage of the minimization process from \cite{HunterPolbergPotyka18}
deals with the uniqueness problem. Among those probability functions that 
minimize the atomic distance, we pick the unique one that minimizes a
sufficiently strong second distance measure.
Here, we will consider again the least-squares distance for stage 2.
\begin{definition}[Two-stage Least-squares Update Operator (2LS)]
The \emph{2LS update operator} $\lsUpdate: \probDists \times \constraints  \rightarrow \probDists \cup \{\bot\}$ is defined by
\begin{equation*}
\lsUpdate(\prob, C) = \begin{cases}
\arg \min_{\prob' \in \lsFirstUpdate(\prob,C)} \sum_{w \in \Omega} (\prob(w) - \prob'(w))^2, & \textit{if $\lsFirstUpdate(\prob,C) \neq \emptyset$} \\
\bot & \, \text{otherwise.}
\end{cases}
\end{equation*}
\end{definition}
Before looking at an example, we note that $\lsUpdate$ is an epistemic update operator as defined in Definition \ref{def_ep_update_function}.
\begin{proposition}
The 2LS update operator is an epistemic update operator.
\end{proposition}

\ifnum\showProofs=1
\begin{proof}
Let us first look at $\lsFirstUpdate(\prob,C) = \arg \min_{\prob' \in \satp(C)} \lsDistance(\prob, \prob')$.
The optimization problem has linear constraints (non-negativity and normalization of $\prob'$ and linear
atomic constraints $C$) and a convex and quadratic objective function. Therefore, $\lsFirstUpdate(\prob,C)$ is convex and closed
and since $\probDists$ is bounded, $\lsFirstUpdate(\prob,C)$ is also bounded and therefore compact.
In particular, $\lsFirstUpdate(\prob,C)$ is non-empty iff $\satp(C) \neq \emptyset$, that is, iff $C$ is satisfiable.

$\lsUpdate(\prob, C) = \arg \min_{\prob' \in \lsFirstUpdate(\prob,C)} \sum_{w \in \Omega} (\prob(w) - \prob'(w))^2$ is defined by minimizing a strictly convex and quadratic
objective function over the compact and convex set $\lsFirstUpdate(\prob,C)$ and therefore has a unique solution whenever
$\lsFirstUpdate(\prob,C) \neq \emptyset$. In particular, since $\lsFirstUpdate(\prob,C) \subseteq \satp(C)$,
the solution must satisfy $C$.
As explained before $\lsFirstUpdate(\prob,C) \neq \emptyset$ iff $C$ is satisfiable. This proves success and
failure. Representation invariance follows from the fact that equivalent constraint sets
$C_1$, $C_2$ yield the same feasible region because $\satp(C_1) = \satp(C_2)$.
Finally, idempotence follows from the fact that $\prob^* = \lsUpdate(\prob, C)$ already satisfies $C$ and has distance
$0$ to itself and therefore is the unique solution of the optimization problem.
\qed
\end{proof}
\fi
\begin{example}
Consider again $\prob_1$ and the constraint $l_1$ from Example \ref{example_atomic_update_not_unique}.
$\prob_6$, shown in Table \ref{tab:exampleAtomicUpdates}, is the unique solution
that minimizes the least-squares distance to $\prob_1$ among those distributions that
minimize the ALS distance to $\prob_1$. That is, $\lsUpdate(\prob_1, \{l_1\}) = \prob_6$.
\end{example}
\begin{example}
As another example, we consider again the scenario from Example \ref{example_implausible_update}
where a one-stage update changed the belief in $B$ in an implausible way.
We get $\lsUpdate(\prob_1, \{\pi(A)=1\}) = \prob_7$ shown in Table \ref{tab:exampleAtomicUpdates}. 
In particular, we have $\prob_7(B) = 0.7 = \prob_1(B)$ as desired.
\end{example}
Intuitively, stage 1 determines which atomic beliefs in arguments have to be changed in order
to satisfy the new constraints. This avoids the counterintuitive behaviour of 
elementwise minimization over the possible worlds, but does not yield a unique solution. Therefore,
stage 2 performs an elementwise minimization over the possible worlds to pick a best solution
among the ones that minimize the change in atomic beliefs. 

\section{Updates over Probability Labellings}

The two-stage minimization process solves our semantical problems, but we are still left with a 
considerable computational problem. This is because we consider probability functions
over possible worlds whose number grows exponentially with the number of arguments in our framework.
However, as illustrated in our previous examples, human reasoning may be guided by atomic beliefs in arguments rather than 
by beliefs in possible worlds. Therefore, a natural question is, what changes semantically 
when considering belief functions over arguments rather than over possible worlds? 
As shown in \cite{potyka2019fragment}, probability functions over possible worlds
can sometimes just be replaced with \emph{probability labellings}
$L: \args \rightarrow [0,1]$ that assign beliefs to atomic arguments directly
without changing the semantics.
We let $\probLabs$ denote the set of all probability labellings.

Formally, probability functions can be related to probability labellings via an equivalence relation \cite{potyka2019fragment}. 
Two probability functions $\prob_1, \prob_2$ are called \emph{atomically equivalent}, denoted as $\prob_1 \equiv \prob_2$, 
iff $\prob_1(A) = \prob_2(A)$ for all $A \in \args$. As usual, 
$[\prob] = \{\prob' \in \probDists \mid \prob' \equiv P\}$ denotes the equivalence class of $\prob$
and $\probDists/\!\equiv \ = \{[\prob] \mid P \in \probDists\}$ denotes the set of all equivalence classes.
As shown in \cite{potyka2019fragmentReport}, there is a one-to-one relationship between $\probDists/\!\equiv$ and $\probLabs$.
\begin{lemma}[\cite{potyka2019fragmentReport}]
\label{lemma_one_to_one_correspondence}
The function $r: \probDists/\!\equiv \ \rightarrow \probLabs$ defined by $r([\prob]) = L_\prob$, where
$L_\prob(A) = \prob(A)$ for all $A \in \args$ is a bijection.   
\end{lemma}
Intuitively, $r$ determines a compact representation of the equivalence class $[\prob]$, namely 
the probability labelling $L_P = r([\prob])$.
Since $r$ is a bijection, every probability labelling can also be related to 
a set of probabity functions $r^{-1}(L_\prob) = [\prob] = \{\prob' \in \probDists \mid \prob' \equiv \prob \}$.
Intuitively, $r^{-1}(L)$ is just the set of probability functions that satify
the atomic beliefs encoded in $L$.
We say that a probability labelling $L$ satisfies a linear atomic constraint
$\sum_{i=1}^n c_i \cdot \pi(A_i) \leq c_0$ iff $\sum_{i=1}^n c_i \cdot L(A_i) \leq c_0$.
The set of probability labellings that satisfy a set of such constraints $C$ is denoted
 by $\satl(C)$.
The following observations from \cite{potyka2019fragmentReport} are helpful to
simplify computational problems by replacing probability functions with probability labellings. 
\begin{lemma}[\cite{potyka2019fragmentReport}]
\label{lemma_constraint_equivalence}
The following statements are equivalent:
\begin{inparaenum}[\itshape (1)\upshape]
	\item P satisfies a linear atomic constraint $l$;
	\item All $\prob' \in [\prob]$ satisfy $l$;
	\item  $L_P = r([\prob])$ satisfies $l$.
\end{inparaenum}
\end{lemma}
For example, in order to decide whether a set of linear atomic constraints $C$ is satisfiable
by a probability function (of exponential size),
we can just check whether it can be satisfied by a probability labelling (of linear size) \cite{potyka2019fragmentReport}. If such a labelling $L$ exists, all probability functions in $r^{-1}(L)$
satisfy $C$. Conversely, if some probability function $\prob$ satisfies $C$, then $L = r([\prob])$ satisfies $C$ as well.

In order to perform updates more efficiently, we could represent epistemic states by
probability labellings. However, we should ask,
what is the relationship between updates over probability functions and updates
over probability labellings? We first note that update operators $\updateOp_W$ that simply minimize the
distance over possible worlds are not necessarily compatible
with atomic equivalence. That is, given a set of linear atomic constraints $C$ and two probability
functions $\prob_1$ and $\prob_2$ such that $\prob_1 \equiv \prob_2$, 
we do not necessarily have $\updateOp_W(\prob_1, C)  \equiv \updateOp_W(\prob_2, C)$.
\begin{example}
Consider $\prob_1$ and $\prob_{8}$ in Table \ref{tab:exampleAtomicUpdates}.
We have $\prob_1(A) = 0.6 = \prob_{8}(A)$ and $\prob_1(B) = 0.7 = \prob_2(B)$, that is,
$\prob_1 \equiv \prob_8$.
Suppose, we update with $C = \{\pi(A) = 0.5	\}$ and update by just minimizing
the least-squares distance to $\prob_1$.
Then $\updateOp_{W}(\prob_1, C) = \prob_{9}$ and $\updateOp_{W}(\prob_{8}, C) = \prob_{10}$, where $\prob_{9}, \prob_{10}$ are 
again shown in Table \ref{tab:exampleAtomicUpdates}.
We have $\prob_{9}(B) = 0.7 \neq 0.65 = \prob_{10}(B)$, that is, $\prob_{9} \not \equiv \prob_{10}$.
\end{example} 
Update operators based on atomic distance measures give us compatibility guarantees that
 we explain in the following proposition.
\begin{proposition}
\label{prop_atomic_compatibility}
Let $\prob_1, \prob_2 \in \probDists$ and let $C \subset \constraints$ be a finite set of linear atomic constraints. If $\prob_1 \equiv \prob_2$, then
\begin{enumerate}
	\item  $\lsDistance(\prob_1,\prob) = \lsDistance(\prob_2,\prob)$ for all $\prob  \in \probDists$,
 \item $\lsFirstUpdate(\prob_1, C) = \lsFirstUpdate(\prob_2, C)$,
 \item $\prob_1' \equiv \prob_2'$ for all $\prob_1', \prob_2' \in \lsFirstUpdate(\prob_1, C)$,
 \item $\lsUpdate(\prob_1, C) \equiv \lsUpdate(\prob_2, C)$.
\end{enumerate}
\end{proposition}
\ifnum\showProofs=1
\begin{proof}
1. Since $\prob_1 \equiv \prob_2$, we have $\prob_1(A) = \prob_2(A)$ for all $A \in \args$.
Hence,
$\lsDistance(\prob_1,\prob) = \sum_{A \in \args} (\prob_1(A) - \prob(A))^2
 = \sum_{A \in \args} (\prob_2(A) - \prob(A))^2 = \lsDistance(\prob_2,\prob)$.

2. By applying item 1, we get 
$\lsFirstUpdate(\prob_1, C) = \arg \min_{\prob' \in \satp(C)} \lsDistance(\prob_1, \prob')
 = \arg \min_{\prob' \in \satp(C)} \lsDistance(\prob_2, \prob') = \lsFirstUpdate(\prob_2, C)$.

3. If $C$ is not satsifiable, $\lsFirstUpdate(\prob_1, C) = \emptyset$ 
and the statement is trivially true, so let us assume
that $C$ is satisfiable.
Let $L_1 = r([\prob_1]), L_1' = r([\prob_1']), L_2' = r([\prob_2'])$ denote the probability labellings corresponding
to $\prob_1, \prob_1', \prob_2'$. It suffices to show that $L_1' = L_2'$. Consider the least-squares distance
$d_\lambda^2(L,L') = \sum_{A \in \args} (L(A) - L'(A))^2$ over probability labellings.
$d_\lambda^2$ is continuous and strictly convex in both arguments.
$\satl(C)$ forms again a compact and convex set. 
Therefore, the minimization problem $\min_{L' \in \satl(C)} \lsDistance(L_1, L')$
has a unique solution $L^*$. For all probability labellings $L$ and corresponding
$\prob \in r^{-1}(L)$, we have
$d_\lambda^2(L_1,L) = \sum_{A \in \args} (L_1(A) - L(A))^2
 = \sum_{A \in \args} (\prob_1(A) - \prob(A))^2 = \lsDistance(\prob_1, \prob)$.
Since $L^*$ is minimal, $d_\lambda^2(L_1,L^*) \leq d_\lambda^2(L_1,L_1')$.
But since $\prob_1'$ minimizes the ALS distance to $\prob_1$, we must also have
$d_\lambda^2(L_1,L_1') =\lsDistance(\prob_1, \prob_1') \leq  \lsDistance(\prob_1, P^*) = d_\lambda^2(L_1,L^*)$. Hence, $d_\lambda^2(L_1,L^*) = d_\lambda^2(L_1,L_1')$ and
 by uniqueness of $L^*$, we must have $L^* = L_1'$.
Analogously, we can show that $L^* = L_2'$ and therefore $L_1' = L_2'$.

4. Note that $\lsUpdate(\prob_1, C) = \arg \min_{\prob' \in \lsFirstUpdate(\prob_1, C)} \sum_{w \in \Omega} (\prob_1(w) - \prob'(w))^2$ and
$\lsUpdate(\prob_2, C) = \arg \min_{\prob' \in \lsFirstUpdate(\prob_2, C)} \sum_{w \in \Omega} (\prob(w) - \prob'(w))^2$.
According to item 2, the feasible regions of both minimization problems are equal.
Since they minimize a different objective function, the minima may be different.
However, according to item 3, all elements in the feasible region are atomically equivalent.
Therefore, the results of both minimization problems must still be atomically equivalent.
\qed
\end{proof}
\fi
Item 1 says that the ALS distance is invariant under atomically equivalent probability functions.
This implies that the updates that minimize the ALS distance are invariant as well (item 2).
As we demonstrated in Example \ref{example_atomic_update_not_unique}, such updates do
not necessarily yield a unique solution. However, when using the ALS distance, we can guarantee that all
solutions are atomically equivalent (item 3). 
This implies that the 2LS update operator is invariant under atomically equivalent 
probability functions in the sense that it yields equivalent results when the
prior probability functions are equivalent (item 4).

\begin{figure}[tb]
	\centering
		\includegraphics[width=0.80\textwidth]{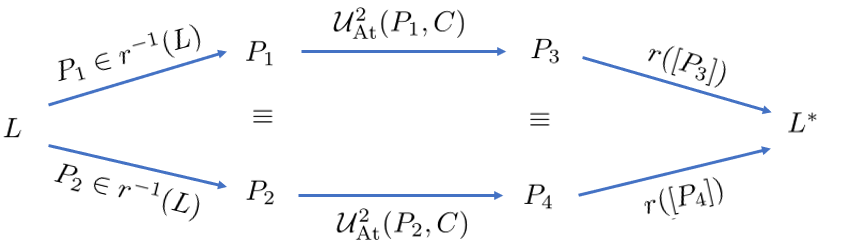}
	\caption{The 2LS update operator $\lsUpdate$ respects atomic equivalence. }
	\label{fig:compatibility_visualization}
\end{figure}

Hence, when updating with respect to linear atomic constraints, there is a well defined
relationship between probability functions and probability labellings.
If we start with an epistemic state represented by a probability labelling $L$,
$L$ can be understood as a compact representation of the set of probability functions $r^{-1}(L)$ that
satisfy the atomic beliefs encoded in $L$.
The 2LS update operator is compatible with this representation.
That is, no matter which probability functions from $r^{-1}(L)$ we choose, an update 
with linear atomic constraints will always lead to the same equivalence class and therefore to
a well defined next probability labelling $L^*$. 
We illustrate this in Figure \ref{fig:compatibility_visualization}.

If we are only interested in atomic beliefs, it would be convenient if we could move directly from $L$ to $L^*$
in Figure \ref{fig:compatibility_visualization} without generating (exponentially large) probability functions in the process.
We can do this indeed in polynomial time for the 2LS update operator.
In order to show this, we first define an update operator on labellings.
\begin{definition}[Least-squares Labelling Update Operator (2LS)]
The \emph{LS labelling update operator} $\lsLabellingUpdate: \probLabs \times \constraints  \rightarrow \probLabs \cup \{\bot\}$ is defined by
\begin{equation*}
\lsLabellingUpdate(L, C)  = \begin{cases}
\arg \min_{L' \in \satl(C)} \sum_{A \in \args} (L(A) - L'(A))^2, & \textit{if $\satl(C) \neq \emptyset$} \\
\bot & \, \text{otherwise.}
\end{cases}
\end{equation*}
\end{definition}
As we explain in the following theorem, 
$\lsLabellingUpdate$ provides us with a direct path from $L$ to $L^*$
and can be computed in polynomial time.
\begin{theorem}
Let $C \subset \constraints$ be a finite and satisfiable set of linear atomic constraints and let  $L  \in \probLabs$.
Then $\lsLabellingUpdate(L, C) = L^*$ is well-defined and can be computed in polynomial time.
Furthermore, $L^* = r([\lsUpdate(\prob, C)])$ for all $\prob \in r^{-1}(L)$.
\end{theorem}

\ifnum\showProofs=1
\begin{proof}
The optimization problem 
$\arg \min_{L' \in \satl(C)} \sum_{A \in \args} (L_1(A) - L'(A))^2$
 that defines $\lsLabellingUpdate(L, C)$ 
has a strictly convex and quadratic objective function and linear constraints.
Therefore, it can be solved in polynomial time, for example, by using the ellipsoid method \cite{kozlov1979polynomial}. In particular, continuity and strict convexity of the objective function and convexity and compactness of the feasible region imply that it has a unique solution.

Now consider 
$
U_{(\mathrm{At},d)}(\prob, C) = \arg \min_{\prob' \in \lsFirstUpdate(\prob, C)} d(\prob, \prob').
$
for some $\prob \in r^{-1}(L)$.
We know from item 4 in Proposition \ref{prop_atomic_compatibility} that 
the solutions for different choices of $\prob$ are all atomically equivalent and therefore
correspond to the same probability labelling.
For one particular choice of $\prob$, we know from item 3 in Proposition \ref{prop_atomic_compatibility} that 
all feasible $\prob' \in \lsFirstUpdate(\prob, C)$ are atomically equivalent as well.
Therefore, they are all mapped to the same probability labelling by $r$.
Similar to the proof of item 3 in Proposition \ref{prop_atomic_compatibility},
we can show that this probability labelling must be $L^*$.
Since $\prob^* \in \lsFirstUpdate(\prob, C)$, it follows $L^* = r([\prob^*])$.
\qed
\end{proof}
\fi
Hence, when we are only interested in atomic beliefs, 
we can use probability labellings to represent epistemic states and use the least-squares labelling 
update operator for updates. Semantically, this is equivalent to regarding epistemic states
as sets of probability functions that satisfy the same atomic beliefs and updating with respect
to the 2LS update operator. The benefit of the labelling representation is that we can perform
updates in polynomial time. 

\section{Application Example}
\label{sec:example}

In this section we come back to the graph in Figure \ref{fig:dialogue2} and analyze a scenario that, 
while being hypothetical, uses the data from an empirical study in \cite{studylink}.
In this study, the user's belief in argument A changed from $0$ to $0.19$ during the dialogue.\footnote{We note that the study data contained examples of dialogues that resulted in a bigger belief change, however, we have chosen this one due to its interesting structure.}. 

The graph in Figure \ref{fig:dialogue2} is generated from an existing dialogue that involved 
an automated dialogue system and a human user.  
Arguments at even depth (starting from $A$) 
are system arguments ($A$, $C$, $G$, $H$, $L$ and $M$), while the ones at odd depth 
are user arguments. The agents take turns in uttering their arguments (starting with $A$), and arguments at the same 
depth are uttered at the same point by a given party. 
We observe that not all user arguments are met with a system response (see arguments $E$ and $K$). 
Despite this fact, the presented arguments have led to a positive change in belief in $A$, contrary to what would be the intuition from 
the classical Dungean approaches. It is possible that if all of the user's counterarguments were addressed, then 
the belief increase would be even more prominent.

We can try to provide an explanation for the belief change observed in \cite{studylink} 
by modeling the reasoning process in our framework.
Let us assume that the constraints representing the user's reasoning demand that the belief 
in an argument is dual to the belief in the average of its attackers. That is, we assume 
$P(X) = 1 - \frac{1}{|Att(X)|} \sum_{Y \in Att(X)} P(Y)$, where $Att(X) = \{Y \in \args \mid (Y,X) \in \attacks\}$). This assumption leads to the following set of constraints:
\begin{equation*}
\begin{split}
 C &= \{ \pi(A) + \pi(B) = 1
, \pi(B) + \pi(C) = 1  
, \pi(D) + \pi(G) = 1 
, \pi(F) + \pi(H) = 1 
,\\&   \pi(C) + 0.33 \pi(D) + 0.34   \pi(E) +  0.33  \pi(F) = 1 
, \pi(C) +  0.5 \pi(I) +  0.5 \pi(J) = 1 
, \\& \pi(H) + \pi(K) = 1 
, \pi(I) + \pi(L) = 1
,  \pi(J) + \pi(M) = 1\}
\end{split}
\end{equation*}


%

\begin{table}[bt]
	\centering
		\begin{tabular}{l|cccccccccccccc}
		 $L$ & $A$ & $B$ & $C$& $D$& $E$& $F$& $G$& $H$& $I$& $J$& $K$& $L$& $M$ \\ 
\hline
 $L_0$ &0	&	1	&	0 	&	1	&	1	&	1	&	0	&	0	&	1	&	1	&	1	&	0	&	0	\\

 $L_1 = \lsLabellingUpdate(L, C \cup \Phi)$ &0.19 & 0.81 & 0.19 	& 0.505 	& 0.975 	& 0.95 	& 0.495	& 0.05	& 0.92	&  0.09	& 0.95	& 0.08	& 0.91 \\
		\end{tabular}
	\caption{Probability labelings before and after the dialogue from Section \ref{sec:example}.}
	\label{tab:casestudy}
\end{table}

Let us further assume that the user initially completely accepts his or her own arguments and completely rejects the system's arguments. This belief state is represented by the labeling $L_0$ shown in Table \ref{tab:casestudy}. 
We now consider a possible persuasion system which, once a given dialogue branch is exhausted, asks the user 
about his or her beliefs in the unattacked arguments. In our case, the user states that he or she believes
$L$, $M$, $E$ and $K$ with the degrees $0.08$, $0.91$, $0.975$ and $0.95$ respectively. This produces constraints 
$\Phi = \{\pi(L) = 0.08, \pi(M) = 0.91, \pi(E) = 0.975, \pi(K) = 0.95\}$. We can use this information along with $C$ to update $L_0$ without asking the user his or her beliefs in all possible arguments. 
The resulting labeling $L_1 = \lsLabellingUpdate(L, C \cup \Phi)$ is shown in Table \ref{tab:casestudy}.

We observe that the belief in $A$ in and $L_0$ and $L_1$ match the expected beliefs $0$ and $0.19$ 
based on the data in \cite{studylink}. 
%
%

\section{Related Work}

There is a large variety of other probabilistic argumentation approaches
\cite{dung2010towards,li2011probabilistic,rienstra2012towards,hunter2014,doder2014probabilistic,polberg2014probabilistic,thimm2017probabilities,KidoO17,rienstra2018probabilistic,ThimmCR18,riveret2018labelling}, which basically differ in the level of detail (e.g., structured or abstract argumentation), in the way how uncertainty
is introduced (e.g. possible worlds correspond to argument interpretations or the graph structure)
and in the nature of uncertainty (e.g., uncertainty about the acceptance state or uncertainty about
the nature of a relation between arguments).

One limitation when restricting to probability labellings is that we cannot compute the probabilities
of complex formulas over arguments anymore without adding further assumptions. 
However, as we demonstrated, we can sometimes do without complex formulas. 
In this context, probability labellings can be seen as an alternative to weighted argumentation
frameworks that also assign a strength value between $0$ and $1$ to arguments \cite{baroni2015automatic,rago2016discontinuity,amgoud2017evaluation,mossakowski2018modular,potyka2018Kr}.
What makes probability labellings an interesting alternative is their well-defined relationship
to probability functions and probability theory.

The problem of adapting an epistemic state with respect to new knowledge has been studied extensively
in the belief revision literature that evolved from the AGM theory developed in \cite{AGM85}.
An up-to-date discussion of the main ideas can be found in \cite{hansson17}.
Our postulates are inspired by AGM postulates. For example, \emph{Success} and \emph{Representation Invariance} can be seen as the 
counterparts of the \emph{Closure} and \emph{Extensionality} postulates in AGM theory. 
The closest relative to our
setting is probably the probabilistic belief change framework from \cite{Kern-Isberner00d}. 
For a discussion of relationships between classical and probabilistic belief changes, see \cite{Kern-Isberner00d} and \cite{KI08}.

Other equivalence relations have been studied in order 
to improve the computational performance of probabilistic reasoning algorithms 
\cite{fischer1996tabl,kern2004combining,finthammer2012using,potyka2016solving}.
However, usually, these equivalence relations are introduced over possible worlds,
not over probability functions. They can be applied to more expressive reasoning formalisms
(they are not restricted to atomic beliefs),
but identifying compact representatives for the corresponding
equivalence classes remains intractable in general \cite{potyka2015concept}.

\section{Conclusions}

We demonstrated that, in the fragment of linear atomic constraints,
it is possible to relate updates over probability labellings
to equivalent updates over classes of probability functions.
This is interesting from a cognitive,  a probabilistic-logical 
and   a computational perspective.
Atomic beliefs are often easier to understand for humans. If we can relate these
beliefs to probability functions, we get a strong foundational basis.
Finally, they can be stored much more compactly and give us polynomial runtime
guarantees.
Our results can probably be generalized to other two-stage update operators. 
However, the building blocks for the two stages have to be chosen carefully in order
to guarantee that the update operator respects atomic equivalence. 
For example, it may not be possible to relate the two-stage update process considered in 
\cite{HunterPolbergPotyka18}, Section 5, to an update operator over probability labellings
 in a meaningful way. 
However, we may be able to construct similar relationships
by replacing the least-squares distance with KL-divergence or more general
classes of distance measures.   
An implementation of our update operator is available 
in the Java library \emph{ProBabble}\footnote{\url{https://sourceforge.net/projects/probabble/}}.


%


\bibliographystyle{splncs03}  

\bibliography{references}

\begin{thebibliography}{10}
\bibitem{studylink}
Anonymous, A.: Argumentation dialogues with participants: An empirical study.
  {To be posted on Arxiv}, {This study has been completed and a report will be
  posted shortly.}
\end{thebibliography}

\end{document}